\pdfoutput=1
\documentclass[10pt]{article}
\PassOptionsToPackage{table,dvipsnames}{xcolor}
\usepackage{leapwing}
\definecolor{cvprblue}{rgb}{0.21,0.49,0.74}

\newcommand{\TODO}[1]{}

\makeatletter
\renewcommand\paragraph{\@startsection{paragraph}{4}{\z@}{1.8ex \@plus .4ex \@minus .2ex}{-1em}{\normalfont\normalsize\bfseries}}
\makeatother

\usepackage{booktabs}
\usepackage{multirow}
\usepackage{xspace}
\newcommand{\method}{JEPA-ICL\xspace}

\usepackage{colortbl}
\definecolor{oursrow}{RGB}{232,241,251}
\definecolor{ablgray}{RGB}{120,120,120}
\newcommand{\best}[1]{\textbf{#1}}
\newcommand{\second}[1]{\underline{#1}}
\newcommand{\abl}[1]{{\color{ablgray}#1}}

\usepackage{pifont}
\newcommand{\yes}{\ding{51}}
\newcommand{\no}{\ding{55}}
\newcommand{\tabgroup}[2]{\multicolumn{#1}{l}{\scriptsize\textit{#2}}}
\definecolor{pvla}{RGB}{86,110,180}    
\definecolor{picl}{RGB}{163,82,204}    
\definecolor{pours}{RGB}{230,126,34}   
\definecolor{bartrack}{RGB}{228,231,238}
\definecolor{dpos}{RGB}{46,140,70}
\definecolor{dneg}{RGB}{200,50,50}
\newlength{\barlen}
\newcommand{\sr}[2]{\makebox[\barlen][l]{\raisebox{0.35ex}{\makebox[0pt][l]{\color{bartrack}\rule{\barlen}{0.55ex}}\color{#1}\rule{\fpeval{#2/100}\barlen}{0.55ex}}}\,\makebox[1.9em][r]{#2}}

\newcommand{\pdot}[1]{\textcolor{#1}{\raisebox{0.1ex}{\scriptsize$\bullet$}}\,}

\usepackage{subcaption,wrapfig}
\providecommand{\ie}{\emph{i.e.}\xspace}

\usepackage[capitalize]{cleveref}
\hypersetup{colorlinks=true,allcolors=LWInk}
\renewcommand{\LWTitle}{JEPA-ICL: Progress-Aligned Latent Plans for In-Context Robotic Manipulation}
\renewcommand{\LWShortTitle}{JEPA-ICL}
\renewcommand{\LWPartnerLogoHeight}{11mm}
\renewcommand{\LWPartnerLogos}{%
  \raisebox{-.30\height}{\includegraphics[width=23mm,height=\LWPartnerLogoHeight,keepaspectratio]{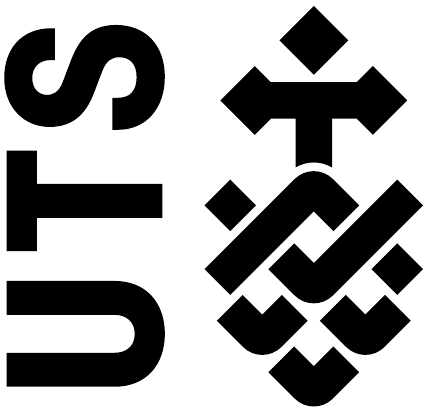}}\hspace{5mm}%
  \raisebox{-.30\height}{\includegraphics[width=12mm,height=\LWPartnerLogoHeight,keepaspectratio]{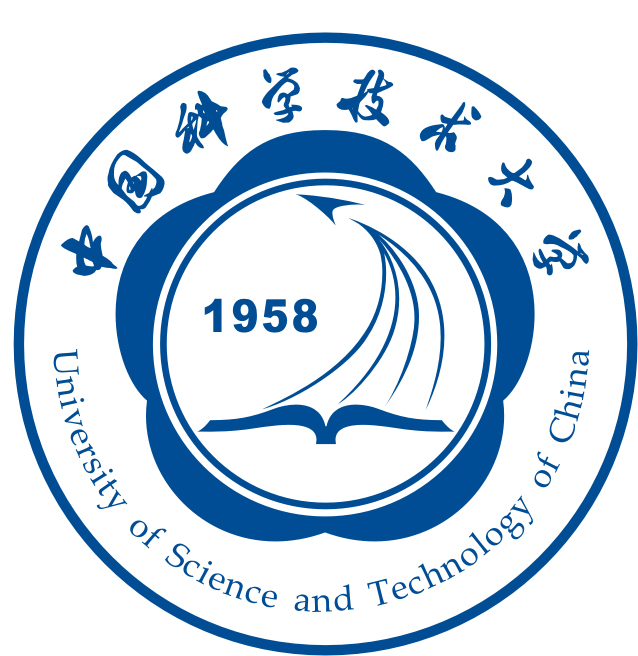}}}
\renewcommand{\LWAuthors}{%
Tingjun Dai\textsuperscript{1*}\quad Jianxiang He\textsuperscript{2*}\quad
Mingfei Han\textsuperscript{3,4*\textdagger}\quad Qihao Yu\textsuperscript{1}\quad Longtao Jiang\textsuperscript{2}\\[3pt]
Changlin Li\textsuperscript{5}\quad Minghao Jin\textsuperscript{4}\quad
Zhihui Li\textsuperscript{4}\quad Xiaojun Chang\textsuperscript{2,3,4}}
\renewcommand{\LWAffiliations}{%
\textsuperscript{1}AAII, UTS\quad \textsuperscript{2}MBZUAI\quad 
\textsuperscript{3}LeapWing AI\quad
\textsuperscript{4}Future Media Computing Lab, USTC\quad \textsuperscript{5}Stanford University}
\renewcommand{\LWNotes}{{\sffamily\color{LWMuted}\textsuperscript{*}Equal contribution\quad\textsuperscript{\textdagger}Project lead}}
\renewcommand{\LWAbstract}{\begingroup\renewenvironment{abstract}{}{}\begin{abstract}
Demonstrations show how a task can be completed, but differences in embodiment, appearance, and execution speed make it difficult to identify the guidance relevant to the robot’s current progress. We introduce JEPA-ICL, a vision-language-action framework that couples predictive planning with demonstration learning. The policy learns to plan by predicting ordered future states in JEPA feature space up to subtask completion, capturing what remains to be accomplished. These plans guide action generation and locate relevant stages in demonstrations, whose JEPA-supervised content features provide progress-aligned context. Learning from this context, in turn, updates the shared policy that produces the plans and actions. This coupling connects anticipating the robot’s own execution with reusing demonstrated experience. On the Imitator Game, JEPA-ICL achieves 80.0\% seen-task success under C15 pretraining, outperforming Zero-WAM by 12.0 percentage points. Without contextual demonstrations, JEPA-ICL outperforms the VLA-JEPA variant without human-video pretraining, improving success rates from 96.1\% to 97.6\% on LIBERO and from 62.9\% to 73.2\% on LIBERO-Plus. With demonstration-based training and contextual guidance at inference, LIBERO-Plus success further increases from 73.2\% to 75.2\%.
\end{abstract}
\endgroup}
\begin{document}
\makeLWTitle
\section{Introduction}
\label{sec:intro}

\ifdefined\LWShortTitle
\begin{wrapfigure}{r}{.49\textwidth}
\vspace{-8pt}\captionsetup{font=footnotesize}
    \centering
    \includegraphics[width=\linewidth]{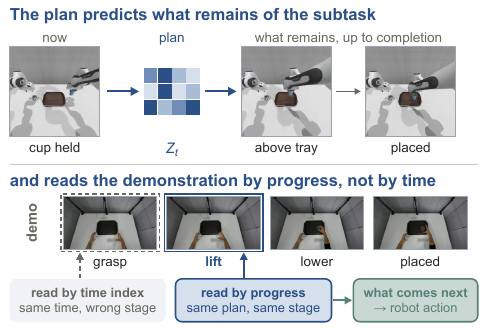}
    \caption{\textbf{A progress-aligned latent plan guides execution and reads demonstrations.}
    Top: from the current observation, the plan $Z_t$ predicts the JEPA features of the checkpoints that remain in the current subtask. Bottom: the same plan locates the demonstration stage at the same progress, where a time index points to the wrong stage; what the demonstration shows next at that stage is fed back to the policy.}
    \label{fig:1}
\vspace{-6pt}\end{wrapfigure}
\else
\begin{figure}[t]
    \centering
    \includegraphics[width=\linewidth]{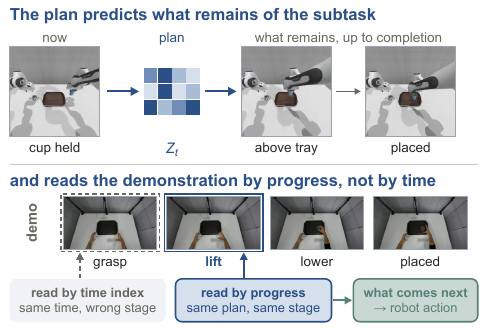}
    \caption{\textbf{A progress-aligned latent plan guides execution and reads demonstrations.}
    Top: from the current observation, the plan $Z_t$ predicts the JEPA features of the checkpoints that remain in the current subtask. Bottom: the same plan locates the demonstration stage at the same progress, where a time index points to the wrong stage; what the demonstration shows next at that stage is fed back to the policy.}
    \label{fig:1}
\end{figure}
\fi

Demonstrations provide more than examples of successful actions: they reveal how a task progresses toward completion. To use this experience, a robot must relate the demonstrated process to what remains to be accomplished in its own execution, even when embodiment, appearance, and execution speed differ. This calls for a representation that supports both anticipating task progress and identifying relevant guidance. Joint embedding predictive architectures (JEPA) offer a natural basis for learning such representations through future feature prediction without requiring image reconstruction.

Recent work has advanced both predictive control and demonstration learning. JEPA-WAM~\cite{jepawam} uses latent transition supervision for action generation, while HOST~\cite{host} connects demonstration localization with future observation prediction. JEDI~\cite{jedi} learns predictive representations for demonstration compression, and MUTEX~\cite{mutex} demonstrates the benefits of learning from complementary task specifications. These advances motivate a more specific question: how can the plan that guides a robot's own actions also organize its use of demonstration experience? Future prediction alone does not establish correspondence across executions, while access to relevant context does not by itself ensure better planning. We study it through the state changes that remain before subtask completion.

We introduce \method, a vision-language-action framework that connects predictive planning and demonstration learning through progress-aligned latent plans. Our central idea is to describe an execution stage by the ordered state changes that remain before the current subtask is complete (\cref{fig:1}). We learn this representation through a plan-conditioned Markov predictor, a latent world model that forecasts future states in JEPA feature space. The resulting plans support action generation and provide progress keys for locating relevant stages in demonstrations. Prediction and demonstration alignment jointly supervise these plans, connecting the ability to anticipate an execution with the ability to learn from another.

Specifically, plan queries from the VLA backbone condition this world model, which starts from the current JEPA state and predicts four ordered checkpoints in the remaining subtask, one shared Markov transition at a time; supervising the intermediate checkpoints as well as completion makes the plan retain the remaining process, not only the endpoint. For demonstration use, the same backbone produces plans for robot observations and demonstration frames, and a learned indexer associates their stages under stage and temporal-consistency supervision. The policy then reads JEPA-supervised content features from a short window around the aligned stage through memory tokens, and the retrieved context is updated during execution so that guidance stays tied to the robot's observed progress.

We evaluate these capabilities separately and in combination. On LIBERO~\cite{libero} and LIBERO-Plus~\cite{liberoplus}, JEPA-ICL achieves 97.6\% and 73.2\% success without contextual demonstrations, respectively; demonstration-based training and contextual guidance at test time further raise LIBERO-Plus success to 75.2\%. On the Imitator Game, JEPA-ICL achieves 80.0\% seen-task success under C15 pretraining, outperforming Zero-WAM~\cite{zerowam} (68.0\%) by 12.0 percentage points. Under C45 pretraining, it reaches 85.5\% unseen-task success after fine-tuning on 10 robot demonstrations per task. Real-robot gear assembly further demonstrates the use of recorded experience: JEPA-ICL achieves 80.0\% success with a deployment-setup demonstration, compared with 40.0\% for the kai0 (\(\pi_{0.5}\)) baseline. Controlled comparisons separate the contributions of latent planning, of progress-aligned retrieval, and of context at test time.

Our contributions are summarized as follows:
\begin{itemize}
    \item \textbf{Markov-supervised latent planning.}
    We learn progress-aware plans through recurrent prediction of ordered future JEPA states within the remaining subtask, supervised at intermediate checkpoints and at subtask completion.

    \item \textbf{Progress-aligned demonstration context.}
    We use latent plans to locate relevant demonstration stages and read separate JEPA-supervised content features through memory tokens, which separates progress correspondence from content retrieval and updates guidance as execution progresses.

    \item \textbf{Joint planning and demonstration learning.}
    We combine future prediction, demonstration alignment, and action supervision to train one shared policy that runs both with and without contextual demonstrations.
\end{itemize}

\section{Related work}
\label{sec:related}

\paragraph{Predictive planning in VLAs.}
VLA models map observations and an instruction to action chunks with a pretrained vision-language backbone~\cite{rt2,openvla,octo,pi0,pi05,groot}. To give them foresight, recent work adds a model of the future: generated frames~\cite{vpp,worldvla,genieenvisioner,dreamvla,dreamzero}, latent actions~\cite{lapa,univla}, or predicted features~\cite{flare,vlajepa,jepawam,juno}. VLA-JEPA~\cite{vlajepa} and JEPA-WAM~\cite{jepawam} are closest to ours, since both supervise a latent plan with frozen V-JEPA features~\cite{vjepa,vjepa2}. Their plan predicts a fixed number of steps ahead, so it says what comes next but not how much of the task remains; progress is instead estimated by a separate model~\cite{gvl} or given to the policy as sub-goals~\cite{stellavla,tpfm,seetraceact}. JEPA-ICL instead predicts ordered future states up to subtask completion. Its latent plan provides a shared interface for action generation and demonstration matching, connecting prediction of the remaining subtask with reuse of demonstrated experience.

\paragraph{Demonstrations as context.}
In-context imitation conditions the policy on a demonstration at test time~\cite{icrt,hist,instantpolicy,synthicl,matchingpolicy,perceiver,vid2robot,contextflow,vivla,ricl,recap,iclrvr,mapvla,zerowam,simpleicl,mimicdroid}. The difficulty is to decide which part of the demonstration matters now. Policies that read it by time index tend to copy or ignore it~\cite{icivla,ravla,stellavla,instantpolicy}, because a long demonstration and a short action chunk rarely advance in step; policies that align by similarity need a hand-crafted metric, such as DTW over actions~\cite{icivla,ravla} or DINO features~\cite{ricl,recap,xicm}; HOST~\cite{host} and Demo-JEPA~\cite{demojepa} learn a dedicated correspondence between demonstration and robot. In these methods the representation used for matching is built apart from the one the policy acts with. \method aligns by progress with the plan it already computes for control, so it needs no hand-crafted metric and no separate correspondence model, only a small projection of the plan, and the stage it reads is the one that fits what remains to be done.

\paragraph{From context back to the plan.}
When the matching representation is fixed, learning to use demonstrations does not change how the policy plans. MUTEX~\cite{mutex} shows that training one policy on several task modalities improves each of them alone, and memory of the policy's own history~\cite{robottt,reflectivevla} improves it within an episode. \method extends this across executions: the action loss under aligned context also updates the backbone that produces the plan.

\section{Method}
\label{sec:method}

\begin{figure*}[t]
    \centering
    \includegraphics[width=\textwidth]{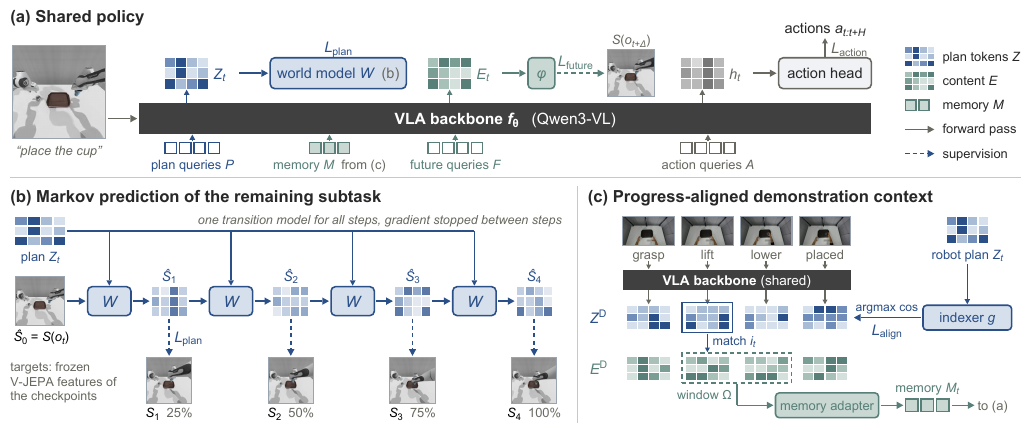}
    \caption{\textbf{One backbone plans, predicts the remaining subtask and reads demonstrations.} (a) Plan, memory, future and action queries share the VLA backbone; plan tokens $Z$ are blue, content features $E$ green; solid arrows are the forward pass, dashed arrows supervision. (b) Conditioned on the plan, the latent world model $W$ predicts the JEPA features of the checkpoints that remain up to subtask completion, one transition at a time. (c) The indexer matches the robot's plan to the plan tokens of the demonstration, the content features around the matched stage are read, and the memory adapter turns them into the memory tokens $M$ the policy attends to.}
    \label{fig:model}
\end{figure*}

\method is a latent-world-model VLA~\cite{vlajepa} whose plan predicts the states that remain until the current subtask is complete. Such a plan describes an execution stage by what is left to do, so it serves two purposes at once: it conditions the robot's actions, and it locates the same stage in a demonstration, whose content is then read back into action generation. \Cref{fig:model} shows the three parts. We first describe the shared policy and its query tokens (\cref{sec:overview}), then how the plan is supervised (\cref{sec:latent_plans}), how it retrieves demonstration context (\cref{sec:demo_context}), and how the whole is trained and run (\cref{sec:joint_learning}).

\subsection{Overview}
\label{sec:overview}
At time $t$ the policy observes camera images $o_t$, an instruction $\ell$ and the proprioceptive state $s_t$, and outputs a chunk of $H$ actions $a_{t:t+H}$. As in VLA-JEPA~\cite{vlajepa}, a vision-language model $f_\theta$ (Qwen3-VL~\cite{qwen3vl}) encodes $o_t$ and $\ell$, and learnable query tokens appended to this sequence carry the outputs of the model (\cref{fig:model}a). The \emph{plan queries} $P$ yield the latent plan $Z_t$; the \emph{future queries} $F$ yield the content features $E_t$; the \emph{action queries} $A$ yield the features $h_t$ that, together with $s_t$, condition a flow-matching action head~\cite{flowmatching,dit}. When a demonstration is available, \emph{memory tokens} $M_t$ computed from it join the same sequence. World states are described by a frozen V-JEPA~2.1 encoder~\cite{vjepa21}, and we write $S(o)$ for the patch features of an image $o$.

The attention mask fixes who may read whom. The plan queries read only the observation and the instruction, so $Z_t$ depends on nothing but the current input. The future and action queries read $Z_t$ and $M_t$, and the action queries do not read the future queries. Every action is thus generated with access to the plan and, when present, to the demonstration context, while the plan is never altered by the context it retrieves. The queries only add tokens to the VLM and read its outputs, so they are not tied to one backbone; \cref{sec:real} instantiates them on a second one, $\pi_{0.5}$~\cite{pi05}.

\subsection{Markov supervision for latent plans}
\label{sec:latent_plans}
\paragraph{Checkpoints in the remaining subtask.}
A target placed a fixed number of steps ahead is tied to time rather than to the task: at the same offset, a slow execution may still be approaching an object that a fast one has already grasped. We therefore define the targets of the plan by the progress that remains in the current subtask. Training episodes are segmented into subtasks, and $e_t$ denotes the end of the subtask that contains $t$. We place $K=4$ checkpoints at $25\%$, $50\%$, $75\%$ and $100\%$ of the remaining interval,
\begin{equation}
    n_k = t + \tfrac{k}{K}\,(e_t - t), \qquad S_k = S(o_{n_k}), \qquad k=1,\dots,K,
    \label{eq:checkpoints}
\end{equation}
so the last target is always the completion of the subtask and the others are the ordered states that lead to it.

\paragraph{Markov prediction.}
A latent world model $W$ predicts the checkpoints one after another. It starts from the current state $\hat S_0 = S(o_t)$ and, at every step, sees only the preceding state and the plan,
\begin{equation}
    \hat S_k = W\big(\hat S_{k-1},\, Z_t\big), \qquad k=1,\dots,K.
    \label{eq:rollout}
\end{equation}
One transition model serves all four steps, and the gradient is stopped between steps so that each transition is trained as a one-step prediction. The plan and the world model are trained to match the checkpoints in feature space,
\begin{equation}
    \mathcal{L}_{\mathrm{plan}} = \sum_{k=1}^{K} \big\lVert w_k \odot (\hat S_k - S_k) \big\rVert_1 ,
    \label{eq:planloss}
\end{equation}
where $w_k$ weights the patches that move within transition $k$ above the static ones (see the appendix).

\paragraph{What the plan carries.}
Because $W$ sees only the current state and $Z_t$, whatever the current frame does not reveal about the rest of the subtask must be carried by the plan, and supervising every checkpoint rather than the endpoint alone asks it to carry the order of the remaining changes, which is what the action head needs to decide the next chunk. Sampling checkpoints relative to subtask completion reduces dependence on a fixed temporal horizon and encourages the plan to represent the remaining process. We use these representations for cross-execution matching in the next section.

\subsection{Progress-aligned demonstration context}
\label{sec:demo_context}
A demonstration helps at the stage that matches the robot's own progress. The time index of a frame cannot locate that stage when the two executions differ in speed, embodiment or layout. The checkpoints of \cref{eq:checkpoints} normalise time within a subtask, which removes the fixed horizon but not the effect of a pause or a correction inside it, so the correspondence between two executions is learned on top of the plan. A demonstration $\mathcal{D}=\{o^{\mathcal{D}}_j\}_{j=1}^{T}$ of the task is passed through the same backbone, without actions or proprioception, to obtain its plans $\{Z^{\mathcal{D}}_j\}$ and content features $\{E^{\mathcal{D}}_j\}$ (\cref{fig:model}c). The robot then locates its stage with the plans and reads the content features there.

\paragraph{Locating the stage.}
A small indexer network $g$ embeds plans, and the matched stage is the demonstration frame whose embedded plan is closest to the robot's,
\begin{equation}
    i_t = \arg\max_{j}\, \cos\big(g(Z_t),\, g(Z^{\mathcal{D}}_j)\big).
    \label{eq:match}
\end{equation}
Frame-level correspondences between two executions are not available, so the indexer is trained with the subtask annotations that already define the plan. The alignment loss is computed on the similarities $s_{ij}=\cos\big(g(Z_i), g(Z^{\mathcal{D}}_j)\big)/\tau$ between sampled frames of a training episode and of a demonstration of the same task, $\mathcal{L}_{\mathrm{align}} = \mathcal{L}_{\mathrm{stage}} + \mathcal{L}_{\mathrm{cycle}} + \omega\,\mathcal{L}_{\mathrm{order}}$. The \emph{stage} term is a contrastive loss whose positives are the demonstration frames annotated with the same subtask as frame $i$; frames without a reviewed label are masked, not used as negatives. The \emph{order} term scores the whole matching path with soft-DTW~\cite{softdtw}, so that matches advance monotonically through the demonstration. The \emph{cycle} term matches the robot softly to the demonstration and back, and penalises the distance in time between the return and the start. The three terms, $\tau$ and $\omega$ are given in the appendix. On the Imitator Game the loss is computed on live plans and trains the plans of both executions as well as the indexer; on LIBERO the demonstration plans are cached and it trains the indexer only.

\paragraph{Content features.}
The plan says where the robot is in the task; the content features say what happens next. The future queries are trained to predict the state a fixed offset $\Delta$ ahead,
\begin{equation}
    \mathcal{L}_{\mathrm{future}} = d\big(\phi(E_t),\, S(o_{t+\Delta})\big),
    \label{eq:future}
\end{equation}
where $\phi$ is a linear projection and $d$ a patch-wise cosine distance. On a demonstration, $E^{\mathcal{D}}_j$ therefore describes the near future of frame $j$ in the same feature space in which the policy predicts its own.

\paragraph{Reading the context.}
The content features in a short window around the matched stage are converted by a memory adapter, a few cross-attention layers with learnable queries, into the memory tokens
\begin{equation}
    M_t = \mathrm{Adapter}\big(\{E^{\mathcal{D}}_{i_t+\delta}\}_{\delta\in\Omega}\big),
    \label{eq:memory}
\end{equation}
which enter the VLM sequence and are read by the future and action queries (\cref{sec:overview}). The policy thus reads from the demonstration what remains to be done at its own stage, not a trajectory to replay. Both the match $i_t$ and the memory $M_t$ are recomputed from every new observation, so the context follows the robot's progress, not elapsed time.

\subsection{Joint policy learning}
\label{sec:joint_learning}
\paragraph{Objective.}
The action head predicts $a_{t:t+H}$ by conditional flow matching. With $a^{\tau} = (1-\tau)\,\epsilon + \tau\, a_{t:t+H}$, $\epsilon\sim\mathcal{N}(0,I)$ and $\tau\sim\mathcal{U}(0,1)$,
\begin{equation}
    \mathcal{L}_{\mathrm{action}} = \mathbb{E}_{\tau,\epsilon}\,\big\lVert v_\psi(a^{\tau}, \tau \mid h_t, s_t) - (a_{t:t+H}-\epsilon)\big\rVert_2^2 ,
    \label{eq:action}
\end{equation}
and the policy is trained with
\begin{equation}
    \mathcal{L} = \mathcal{L}_{\mathrm{action}} + \beta\,\mathcal{L}_{\mathrm{plan}} + \gamma\,\mathcal{L}_{\mathrm{future}} + \lambda\,\mathcal{L}_{\mathrm{align}} .
    \label{eq:total}
\end{equation}
The plan queries are trained by $\mathcal{L}_{\mathrm{plan}}$ through the world model and, on live plans, by $\mathcal{L}_{\mathrm{align}}$ through the indexer; the future queries by $\mathcal{L}_{\mathrm{future}}$. The action loss trains the action head, the memory adapter and, through $h_t$ and $M_t$, the backbone that also produces the plan, which closes the path from plan to context and back to plan.

\paragraph{Training procedure.}
The world model is initialised in a first stage that pretrains $W$ on the training episodes together with a teacher that compresses the true checkpoints into a plan. The policy’s plan is first regressed onto the teacher plan during a short warm-up, after which the teacher-regression loss is replaced by $\mathcal{L}_{\mathrm{plan}}$ (see the appendix). During policy training, a demonstration of the same task from a different episode is attached to most samples, and $M_t$ is set to zero for the rest so that the policy remains usable without one. The plans and content features of the demonstrations are computed by a copy of the policy that is refreshed periodically, which keeps the keys stable while the policy changes.

\paragraph{Execution.}
At test time the encoder, the world model and the teacher are discarded. Without a demonstration the policy runs the VLM once per chunk with $M_t=0$, so the plan adds only its query tokens to the inference cost. With a demonstration, its frames are encoded once and their plans and content features are cached; at every chunk the policy matches its current plan against the cached plans (\cref{eq:match}), reads the window around the match (\cref{eq:memory}) and generates the chunk. The plan queries precede the memory tokens and do not read them, so $Z_t$ is read out first and the rest of the sequence is processed afterwards; the $\pi_{0.5}$ instance reuses the cached prefix, so the observation is encoded once.

\section{Experiments}
\label{sec:experiments}

We evaluate JEPA-ICL on the Imitator Game, LIBERO, and LIBERO-Plus in simulation, and on a real bimanual robot. The experiments compare policies with and without the predictive plan component and examine performance under different demonstration-context conditions.

\begin{table*}[t]
\centering
\footnotesize
\setlength{\tabcolsep}{3.4pt}
\renewcommand{\arraystretch}{1.0}
\caption{
\textbf{Main results on the Imitator Game}~\cite{imitatorgame}
(simulation, \%). C15 and C45: pretraining on 15 or 45 tasks; each entry averages five seen or unseen tasks over the four imitation levels L0--L3. \emph{P+FT}: the pretrained policy is fine-tuned on 10 robot demonstrations per unseen task. \emph{Sub-SR}: the benchmark-defined subtask progress, the fraction of a task's stages whose sub-reward reaches its threshold within the episode. Baselines are aggregated from the official per-scale leaderboard data. Our row is the full \method, given the human video. Colours mark the paradigm.
}
\label{tab:imitator}
\begin{tabular}{l l c c c c c c c c}
\toprule
\multirow{3}{*}{Method}
& \multirow{3}{*}{Paradigm}
& \multicolumn{4}{c}{C15: 15-task pretraining}
& \multicolumn{4}{c}{C45: 45-task pretraining} \\
\cmidrule(lr){3-6}\cmidrule(lr){7-10}
& & \multicolumn{2}{c}{Seen}
  & \multicolumn{2}{c}{Unseen (P+FT)}
  & \multicolumn{2}{c}{Seen}
  & \multicolumn{2}{c}{Unseen (P+FT)} \\
\cmidrule(lr){3-4}\cmidrule(lr){5-6}
\cmidrule(lr){7-8}\cmidrule(lr){9-10}
& & SR & Sub-SR & SR & Sub-SR
  & SR & Sub-SR & SR & Sub-SR \\
\midrule
\pdot{pvla}OpenVLA~\cite{openvla}
& VLA
& \sr{pvla}{39.0} & \sr{pvla}{69.6}
& \sr{pvla}{19.0} & \sr{pvla}{40.9}
& \sr{pvla}{22.0} & \sr{pvla}{52.2}
& \sr{pvla}{22.5} & \sr{pvla}{47.8} \\

\pdot{pvla}RDT-1B~\cite{rdt}
& VLA
& \sr{pvla}{37.0} & \sr{pvla}{64.0}
& \sr{pvla}{19.0} & \sr{pvla}{43.5}
& \sr{pvla}{41.0} & \sr{pvla}{73.4}
& \sr{pvla}{24.5} & \sr{pvla}{43.6} \\

\pdot{pvla}GR00T-N1.6~\cite{groot}
& VLA
& \sr{pvla}{29.5} & \sr{pvla}{74.9}
& \sr{pvla}{57.5} & \sr{pvla}{74.8}
& \sr{pvla}{74.5} & \sr{pvla}{91.6}
& \sr{pvla}{73.5} & \sr{pvla}{81.7} \\

\pdot{picl}Zero-WAM~\cite{zerowam}
& ICL
& \sr{picl}{68.0} & \sr{picl}{91.3}
& \sr{picl}{77.0} & \sr{picl}{85.7}
& \sr{picl}{68.0} & \sr{picl}{91.2}
& \sr{picl}{84.5} & \sr{picl}{91.4} \\
\midrule

\rowcolor{oursrow}
\pdot{pours}\textbf{\method (ours)}
& ICL
& \sr{pours}{80.0} & \sr{pours}{91.1}
& \sr{pours}{79.0} & \sr{pours}{87.4}
& \sr{pours}{81.0} & \sr{pours}{91.8}
& \sr{pours}{85.5} & \sr{pours}{92.6} \\
\bottomrule
\end{tabular}
\end{table*}

\begin{table}[t]
\centering
\footnotesize
\setlength{\tabcolsep}{2.9pt}
\renewcommand{\arraystretch}{1.0}
\caption{\textbf{Comparison on LIBERO}~\cite{libero}. Success rate (\%) on the four suites (Long $=$ LIBERO-10) and their average. Our policy runs without a demonstration in context. Baseline numbers are taken from~\cite{vlajepa}; $^\dagger$its variant without human-video pretraining. \textbf{Bold} / \underline{underline}: best / second best.}
\label{tab:libero}
\begin{tabular}{lccccc}
\toprule
\multirow{2}{*}{Method} & \multicolumn{4}{c}{LIBERO suite} & \multirow{2}{*}{Avg.} \\
\cmidrule(lr){2-5}
 & Spatial & Object & Goal & Long & \\
\midrule
\tabgroup{6}{Robot-data pretrained VLAs} \\
$\pi_0$~\cite{pi0}                  & 96.8 & \second{98.8} & 95.8 & 85.2 & 94.2 \\
$\pi_0$-FAST~\cite{fast}            & 96.4 & 96.8 & 88.6 & 60.2 & 85.5 \\
OpenVLA-OFT~\cite{openvlaoft}       & 97.6 & 98.4 & \second{97.9} & \second{94.5} & \second{97.1} \\
GR00T N1~\cite{groot}               & 94.4 & 97.6 & 93.0 & 90.6 & 93.9 \\
$\pi_{0.5}$~\cite{pi05}             & \best{98.8} & 98.2 & \best{98.0} & 92.4 & 96.9 \\
\midrule
\tabgroup{6}{Latent-action \& human-video VLAs} \\
LAPA~\cite{lapa}                    & 73.8 & 74.6 & 58.8 & 55.4 & 65.7 \\
UniVLA~\cite{univla}                & 96.5 & 96.8 & 95.6 & 92.0 & 95.2 \\
villa-X~\cite{villax}               & 97.5 & 97.0 & 91.5 & 74.5 & 90.1 \\
\midrule
\tabgroup{6}{Future-prediction \& world-model VLAs} \\
CoT-VLA~\cite{cotvla}               & 87.5 & 91.6 & 87.6 & 69.0 & 83.9 \\
WorldVLA~\cite{worldvla}            & 87.6 & 96.2 & 83.4 & 60.0 & 81.8 \\
VLA-JEPA$^\dagger$~\cite{vlajepa} & 94.8 & \best{99.6} & 95.8 & 94.0 & 96.1 \\
\midrule
\rowcolor{oursrow}
\textbf{\method (ours)}             & \best{98.8} & 97.6 & \best{98.0} & \best{95.8} & \best{97.6} \\
\bottomrule
\end{tabular}
\end{table}

\begin{table*}[t]
\centering
\footnotesize
\setlength{\tabcolsep}{6pt}
\renewcommand{\arraystretch}{1.0}
\caption{\textbf{Comparison on LIBERO-Plus}~\cite{liberoplus}. Models are trained on the original LIBERO data and evaluated zero-shot under seven perturbation dimensions; each dimension covers all four LIBERO suites. Top: per-dimension numbers reported in~\cite{vlajepa} ($^\dagger$its variant without human-video pretraining). Bottom: our policy without (\no) and with (\yes) one demonstration in context; the \emph{Demo} column marks whether a demonstration is given at test time, and the second row is the in-context checkpoint, trained for 10K further updates with demonstrations. Avg.\ is the macro average over the seven dimensions, recomputed for every row. \textbf{Bold} / \underline{underline}: best / second best. Per-suite results are in the appendix.}
\label{tab:liberoplus}
\begin{tabular}{lc ccccccc c}
\toprule
\multirow{2}{*}{Method} & \multirow{2}{*}{Demo} & \multicolumn{7}{c}{Perturbation dimension} & \multirow{2}{*}{Avg.} \\
\cmidrule(lr){3-9}
 & & Camera & Robot & Language & Light & Background & Noise & Layout & \\
\midrule
\tabgroup{10}{Reported in~\cite{vlajepa}} \\
UniVLA~\cite{univla}                & \no & 1.8  & 46.2 & 69.6 & 69.0 & 81.0 & 21.2 & 31.9 & 45.8 \\
OpenVLA-OFT~\cite{openvlaoft}       & \no & 56.4 & 31.9 & 79.5 & 88.7 & \second{93.3} & \second{75.8} & 74.2 & 71.4 \\
$\pi_0$~\cite{pi0}                  & \no & 13.8 & 6.0  & 58.8 & 85.0 & 81.4 & \best{79.0} & 68.9 & 56.1 \\
$\pi_0$-FAST~\cite{fast}            & \no & \best{65.1} & 21.6 & 61.0 & 73.2 & 73.2 & 74.4 & 68.8 & 62.5 \\
WorldVLA~\cite{worldvla}            & \no & 0.1  & 27.9 & 41.6 & 43.7 & 17.1 & 10.9 & 38.0 & 25.6 \\
VLA-JEPA$^\dagger$~\cite{vlajepa} & \no & 40.3 & \best{55.7} & 72.9 & 88.2 & 70.5 & 38.2 & \second{74.6} & 62.9 \\
\midrule
\rowcolor{oursrow}
\textbf{\method (ours)}      & \no & 55.7 & 49.7 & \second{80.6} & \second{92.6} & 93.1 & 66.5 & 74.2 & \second{73.2} \\
\rowcolor{oursrow}
\textbf{\method (ours)} & \yes
& \second{57.4} & \second{51.1} & \best{82.4} & \best{94.9}
& \best{93.4} & 69.8 & \best{77.2} & \best{75.2} \\
\bottomrule
\end{tabular}
\end{table*}

\begin{table}[t]
\centering
\footnotesize
\setlength{\tabcolsep}{4pt}
\renewcommand{\arraystretch}{1.08}
\caption{\textbf{Real-world gear assembly in a parallel setup.} Same $\pi_{0.5}$ checkpoint, same $100$ demonstrations from the training setup, deployed in the parallel setup without fine-tuning; $20$ trials per row. The three \method rows are one policy: with no demonstration, with one recorded in the training setup, and with one recorded in the deployment setup.}
\label{tab:real}
\begin{tabular}{llc}
\toprule
Method & Demonstration at test time & SR (\%) \\
\midrule
kai0 ($\pi_{0.5}$)~\cite{kai0,pi05} & none & 40.0 \;(8/20) \\
\midrule
\method (ours)              & none                           & 60.0 \;(12/20) \\
\method (ours)              & task only (training setup)     & 65.0 \;(13/20) \\
\rowcolor{oursrow}
\method (ours)              & task + deployment setup        & \textbf{80.0} \;(16/20) \\
\bottomrule
\end{tabular}
\end{table}

\subsection{Setup}
\label{sec:setup}
\paragraph{Benchmarks and protocols.}
\textbf{The Imitator Game}~\cite{imitatorgame} asks a robot to perform a task shown in a human video. Policies are pretrained on the paired human--robot episodes of IG-10K at two scales (C15 and C45: $15$ or $45$ pretraining tasks) and evaluated in simulation on five seen and five unseen tasks. Each task is tested at four imitation levels (L0--L3) that move the robot's scene away from the demonstration, from the same scene to objects that serve the same intent through a different affordance. Seen tasks are evaluated with the pretrained policy; unseen tasks are held out from pretraining and evaluated under the \emph{P+FT} protocol, which first fine-tunes the policy on $10$ robot demonstrations per task: adaptation with limited data, not zero-shot transfer. In both cases our policy receives the human video as context at test time. \textbf{LIBERO}~\cite{libero} has four suites of ten tasks (Spatial, Object, Goal and the long-horizon LIBERO-10) on a single-arm Franka. \textbf{LIBERO-Plus}~\cite{liberoplus} perturbs the same tasks along seven dimensions (camera, robot initial state, language, lighting, background, sensor noise, layout); following its protocol, models are trained on the original LIBERO data only and evaluated zero-shot. On both, the policy runs without a demonstration, except for the row of \cref{tab:liberoplus} that gives it one from a different training episode.

The metric is the task success rate (SR); the Imitator Game adds its subtask progress (Sub-SR, defined in \cref{tab:imitator}), and LIBERO-Plus is summarised by the macro average over its seven dimensions. Baseline numbers in \cref{tab:imitator,tab:libero,tab:liberoplus} are those reported by the benchmarks or by VLA-JEPA~\cite{vlajepa}, except the LIBERO-Plus averages, which we recompute as this macro average for every row; they compare complete systems, and gains are attributed to the plan or to the demonstration only by the within-method comparisons of \cref{fig:abl} and \cref{tab:real}. Our simulation results follow each benchmark's official protocol and come from one training run evaluated once with fixed seeds, so we report no variance.

\paragraph{Implementation.}
On LIBERO and LIBERO-Plus we build on StarVLA~\cite{starvla} with Qwen3-VL-2B~\cite{qwen3vl} as the backbone and a frozen V-JEPA~2.1 ViT-L~\cite{vjepa21} as the target encoder; the policy is trained on all four suites for $30$K updates without demonstrations and, for the in-context model, $10$K further updates with a demonstration attached to most samples, at a global batch of $256$. On the Imitator Game and the real robot we instantiate the same queries on $\pi_{0.5}$~\cite{pi05}, so the method is tested on two backbones. All instances use $K=4$, $16$ plan queries, $\Delta=16$, $H=16$, $\beta=0.02$, $\gamma=0.1$ and $\lambda=0.05$; one action chunk of the Qwen3-VL instance takes about $0.13$\,s on an A100 GPU without a demonstration. Further implementation details are in the appendix.

\subsection{Learning from a human video in context}
\label{sec:imitator_results}
\paragraph{On seen tasks, \method has the highest success at both scales.}
The baselines in \cref{tab:imitator} receive the task as a caption; Zero-WAM~\cite{zerowam} and \method receive the human video. Pretrained on $15$ tasks, \method succeeds on $80.0\%$ of the seen-task episodes, $+12.0$ points over Zero-WAM, the next best. Its Sub-SR, $91.1\%$, is on par with Zero-WAM ($91.3\%$). Pretrained on $45$ tasks, it reaches $81.0\%$ SR and $91.8\%$ Sub-SR, $+6.5$ points over GR00T-N1.6 and $+13.0$ over Zero-WAM.

\paragraph{Adaptation to held-out tasks.}
After fine-tuning on $10$ demonstrations per task, the C15 policy reaches $79.0\%$ SR and $87.4\%$ Sub-SR on the unseen tasks, above Zero-WAM ($77.0\%$) and GR00T-N1.6 ($57.5\%$). Once fine-tuned on a task, the policy gains somewhat less from the video than on the seen tasks ($+6.5$ against $+8.0$ points over the same policy without it, \cref{fig:abl}b). The C45 policy reaches $85.5\%$ SR and $92.6\%$ Sub-SR, above Zero-WAM ($84.5\%$), with the highest Sub-SR in the table. \Cref{fig:abl}b separates the effect of the video from how it is read.

\subsection{Execution and robustness on LIBERO}
\label{sec:execution_robustness}
\paragraph{\method has the highest average on LIBERO among the compared methods.}
Without a demonstration, \method reaches the highest average success rate in \cref{tab:libero}, $97.6\%$ against $97.1\%$ for OpenVLA-OFT, $96.9\%$ for $\pi_{0.5}$ and $96.1\%$ for VLA-JEPA without human-video pretraining, the model it extends. It is best or tied best on Spatial, Goal and the long-horizon LIBERO-10, where its margin is largest ($95.8$ against $94.5$ and $94.0$), consistent with a plan of what remains in the subtask adding most where tasks have the most subtasks. On Object it trails four baselines, VLA-JEPA by the most ($97.6$ against $99.6$). The benchmark is near saturation; \cref{fig:abl}a attributes $0.8$ points of the average to the plan.

\paragraph{The gap widens under perturbation.}
On LIBERO-Plus the same policy averages $73.2\%$, $+1.8$ points over OpenVLA-OFT and $+10.3$ over VLA-JEPA (\cref{tab:liberoplus}). It is the best model in the table under language and lighting perturbations, on par with the best under background and layout, and gains most over VLA-JEPA under camera, noise and background changes ($+15.4$, $+28.3$ and $+22.6$). These are comparisons of complete systems; on the long-horizon suite, the one on which this control was run, removing the plan from the same policy lowers success from $65.3\%$ to $55.8\%$ (appendix). Two dimensions remain weak: robot initial state ($49.7$ against $55.7$ for VLA-JEPA) and sensor noise ($66.5$, below $\pi_0$ and OpenVLA-OFT).

\paragraph{With a demonstration, the in-context model is higher in all seven dimensions.}
The last row of \cref{tab:liberoplus} is the in-context model, trained for $10$K further updates with demonstrations and given one demonstration recorded in the unperturbed training scene. It receives more input than the other rows, and its difference from the row above combines training with demonstrations and reading one at test time, which \cref{fig:abl}a separates. The average rises to $75.2\%$ and every dimension is higher, by the most under noise, layout and lighting ($+3.3$, $+3.0$ and $+2.3$) and by the least under background ($+0.3$).

\subsection{Real-world deployment}
\label{sec:real}
\begin{figure*}[t]
    \centering
    \includegraphics[width=\textwidth]{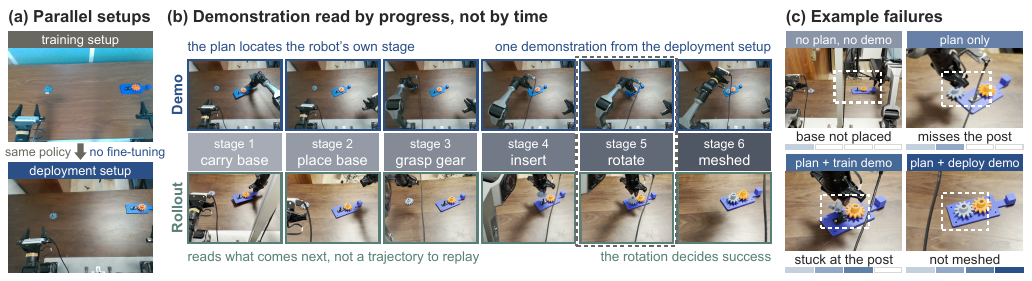}
    \caption{\textbf{A demonstration from the deployment setup is read stage by stage.} (a) Top camera of the two setups: same robot model, different table, lighting and background. (b) The demonstration recorded in the deployment setup and a rollout with it, at matched stages; the dashed box marks the corrective rotation that decides success. (c) The last frame of one failed trial per setting, as examples and not as counts; the bar marks the stage of failure (place the base, align the gear, insert, mesh).}
    \label{fig:real}
\end{figure*}

\paragraph{Setup.}
Two bimanual AgileX Piper robots of the same model, each observed by a top camera and two wrist cameras, stand in two setups side by side (\cref{fig:real}a): the training setup, and a parallel deployment setup with a different table, lighting and background. The task is a two-arm gear assembly: one arm places a base carrying a post and an orange gear at the centre of the workspace; the other inserts a grey gear on the post so that the teeth mesh, and rotates it in place when they do not. Whether this rotation is finished can only be judged from the image, not from the gripper, so the task tests whether stages are told apart by what remains to be done. We collect $100$ demonstrations by teleoperation in the training setup and annotate their subtasks. The policy is the $\pi_{0.5}$ instance of \method through the kai0 implementation~\cite{kai0} of $\pi_{0.5}$; it and the kai0 baseline start from the same checkpoint and are fine-tuned with all parameters on the same data for $20$K updates at a batch of $256$. We evaluate in the deployment setup without fine-tuning; no training data was collected there, and the in-context rows receive one demonstration recorded there. Each setting runs $20$ trials with the same initial configurations; a trial succeeds when the gears mesh, with no time limit and no intervention.

\paragraph{The plan transfers to the parallel setup.}
In the deployment setup, fine-tuned kai0 succeeds in $8$ of $20$ trials, and the plan alone raises this to $12$ (\cref{tab:real}), at the cost of its query tokens only. We attribute this to what the plan predicts: what remains of the current subtask, not the table, lighting or background that changed between the setups.

\paragraph{In this task, a demonstration from the deployment setup helps more than one from the training setup.}
With a demonstration from the training setup, the same policy reaches $13$ of $20$, one trial more than with none and within the noise of $20$ trials: it learned the task from the same $100$ episodes, so seeing the task once more may add little it does not already know. With a demonstration recorded in the deployment setup, the same policy and mechanism reach $16$ of $20$, twice the $8$ of the baseline, without updating a parameter (\cref{fig:real}b). The two demonstrations show the same task on the same robot model and differ in where they were recorded, which suggests that what the policy reads from the second is the deployment setup at each stage.

\paragraph{Both parts carry over to a physical robot.}
In the failed trials of \cref{fig:real}c, one per setting rather than counts, the failure moves later as the plan and the demonstration are added: the baseline stalls before the base is placed, the plan brings the gear to the post, a demonstration of the task gets it onto the post, and with the deployment-setup demonstration what remains is the final mesh.

\subsection{Analysis}
\label{sec:analysis}

\ifdefined\LWShortTitle
\begin{figure}[t]
\centering
\captionsetup{font=footnotesize,skip=5pt}
\begin{minipage}[t]{.45\linewidth}
\vspace{0pt}\centering
\includegraphics[width=\linewidth]{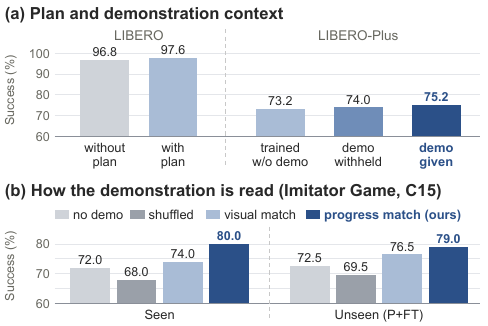}
\caption{\textbf{Latent planning and progress alignment.}
(a) LIBERO compares policies with/without a plan. LIBERO-Plus compares training without context against context training, with the demonstration withheld or provided at test time.
(b) Imitator Game (C15) compares no demonstration, shuffled demonstrations, visual-similarity retrieval, and progress-aligned retrieval.}
\label{fig:abl}
\end{minipage}\hfill
\begin{minipage}[t]{.525\linewidth}
\vspace{0pt}\centering
\includegraphics[width=\linewidth]{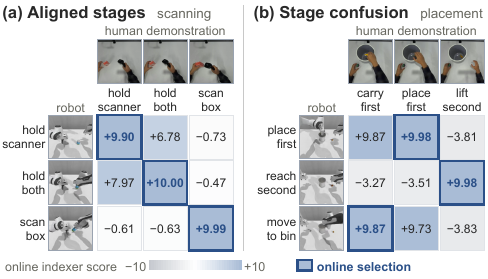}
\caption{\textbf{Cross-embodiment stage matching and failure.}
Online indexer scores (cosine similarity divided by temperature $0.1$): robot plans in rows, three human-demonstration stages in columns. Outlined cells are selected online.
(a) Scanning: retrieval follows the robot's progress.
(b) Two-object placement: after the robot switches to the second object, retrieval returns to the first object's segment.}
\label{fig:ana}
\end{minipage}
\end{figure}
\else
\begin{figure}[t]
    \centering
    \includegraphics[width=\linewidth]{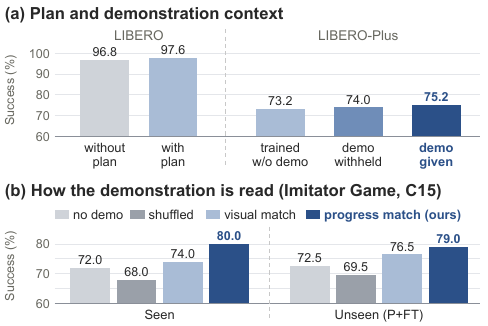}
    \caption{\textbf{Effectiveness of latent planning and progress alignment.}
(a) On LIBERO, we compare policies with and without the latent plan.
On LIBERO-Plus, we compare a policy trained and evaluated without context against a context-trained checkpoint evaluated with the demonstration either withheld or provided.
(b) On the Imitator Game under C15 pretraining, we evaluate progress alignment by comparing the same policy with no demonstration, temporally shuffled demonstrations, visual-similarity retrieval, and progress-aligned retrieval (ours).}
    \label{fig:abl}
\end{figure}

\begin{figure}[t]
    \centering
    \includegraphics[width=\linewidth]{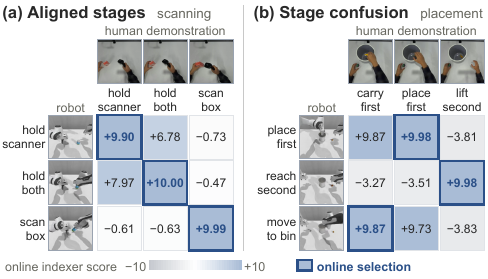}
    \caption{\textbf{The plan locates matching stages across embodiments, though not always.} Online indexer scores (cosine similarity over the temperature $0.1$) between the robot's plan (rows) and three stages of the human demonstration (columns); outlined cells are selected online. (a) Scanning: the selection follows the robot. (b) Two-object placement: once the robot turns to the second object, retrieval returns to the segment of the first.}
    \label{fig:ana}
\end{figure}
\fi

\paragraph{Training with context and reading it at test time are measured separately.}
On LIBERO (\cref{fig:abl}a), adding the plan to the same policy, which keeps its world model in both variants, raises the average from $96.8\%$ to $97.6\%$. On LIBERO-Plus, we compare two checkpoints under three evaluation conditions: the checkpoint trained without context is evaluated without a demonstration ($73.2\%$), while the context-trained checkpoint, which receives $10$K further updates, is evaluated with the demonstration either withheld ($74.0\%$) or provided ($75.2\%$). The context-trained checkpoint is $0.8$ points higher even when no demonstration is given; this difference includes its $10$K additional updates and is consistent with the action loss under aligned context also updating the shared backbone. With the checkpoint fixed, reading a demonstration at test time adds $1.2$ points.

\paragraph{Progress-aligned retrieval yields the largest gains from demonstration context.}
\Cref{fig:abl}b varies how the human video is read on C15 with the same policy. Shuffling its frames in time, so that the retrieved window corresponds to no stage, lowers success below the policy given no video ($-4.0$ points on seen tasks, $-3.0$ on held-out tasks after P+FT); locating the stage by visual similarity recovers part of the gain ($+2.0$ and $+4.0$ over no video), and locating it by the plan gives the largest ($+8.0$ and $+6.5$). A demonstration is thus not a prior that helps whenever present: read at the wrong stage it misleads, and how its stage is located decides how much it helps.

\paragraph{Stage correspondence and limitations.}
Figure~\ref{fig:ana} examines retrieval after C15 task adaptation. In scanning, the selected human frames follow the robot from holding the scanner to bringing both objects together, illustrating how latent plans support stage correspondence across embodiment and appearance differences. In two-object placement, retrieval returns to the first object's demonstration segment after the robot turns to the second, illustrating possible stage confusion between repeated operations.

\section{Conclusion}
\label{sec:conclusion}
We presented JEPA-ICL, a VLA framework that couples predictive planning with demonstration learning. Supervised through recurrent prediction of future JEPA states up to subtask completion, its latent plans guide actions and retrieve demonstration context, while learning from this context updates the shared policy. Experiments on the Imitator Game show the advantage of progress-based retrieval over visual matching, complemented by LIBERO, LIBERO-Plus, and real-robot evaluations with and without contextual demonstrations. These findings suggest that anticipating what remains to be accomplished can provide a shared basis for guiding a robot’s own execution and organizing its use of demonstrated experience. Two questions remain open: our ablations test the plan as a whole rather than its design choices (intermediate checkpoints, recurrent transition), and we do not measure how training with context changes the plan itself.

\section*{LLM Usage Statement}
Large language models were used in three ways in this work; the authors take full responsibility for all content.

\paragraph{Subtask annotation.}
The subtask boundaries that define the progress targets of the plan were produced semi-automatically. Candidate boundaries were taken from gripper open and close events in the training episodes; GPT-5.6 (OpenAI) was then prompted with frames and the task instruction to label the segments and to place the boundaries that the gripper state cannot mark, such as the end of the corrective rotation in the real-robot task. Every segmentation was reviewed and, where necessary, corrected by the authors before it was used for training. No other part of the training data or the evaluation involved an LLM.

\paragraph{Code.}
Claude (Anthropic) assisted in writing the training, evaluation and data-processing code. The model design and the training recipe were specified by the authors, and all code was reviewed, tested and run by the authors.

\paragraph{Writing and literature search.}
Claude (Anthropic) was used as a writing assistant for this manuscript: to edit and condense text drafted by the authors in Sections 2 to 4 and the appendix, to typeset tables, and to locate and check bibliographic entries. The scientific claims, the experimental design, all results and their interpretation are the authors' own; every statement produced with assistance was checked by the authors against the code, the data and the cited papers.

{
    \small
    \bibliographystyle{ieeenat_fullname}
    \bibliography{main}
}

\clearpage
\appendix
\section{Implementation details of the progress-aware plan}
\label{sec:suppl_impl}

\paragraph{Targets.}
On LIBERO, the $1{,}693$ training episodes of the four suites carry fine-subtask segments, and a boundary is anchored on a gripper open or close event wherever the action stream confirms one. The $K=4$ checkpoints (\cref{sec:latent_plans}) are rounded to frames and repeat the end frame when fewer than $K$ frames remain. Each target is encoded as a pair state $S(o_{n_k-1}, o_{n_k})$, the encoder features of the node frame and the frame before it, so that it also carries the local motion at that point; the current state is $S(o_t,o_t)$. Features have $24\times24$ patches of dimension $1024$.

\paragraph{Loss.}
Most patches of a manipulation scene are static, so the $L_1$ plan loss of \cref{sec:latent_plans} is computed separately on the patches that move within each pair (detected by frame differencing) and on the remaining patches, weighted $0.8$ and $0.2$; it falls back to the dense average when either region is empty. The world-model rollout stops the gradient between recurrent steps.

\paragraph{World model and its pretraining.}
The world model is an eight-layer transformer decoder over the patch grid with shifted-window self-attention and cross-attention to the plan ($16\times512$). It is pretrained on the training episodes together with a two-layer resampler that compresses the $K$ target states into a teacher plan, so that the plan is the only channel for future information. In policy training, the plan projection is first regressed onto the teacher plan for $1{,}250$ updates (smooth $L_1$, weight $0.1$); afterwards the teacher is removed and the plan is trained through $\mathcal{L}_{\mathrm{plan}}$. The world model is kept fixed until update $2{,}500$ and then trained jointly; from update $5{,}000$ an additional term of the same form, with the gradient to the plan stopped, keeps it accurate. $\beta$ is annealed linearly from $0.02$ to $0.005$ between updates $5{,}000$ and $30{,}000$ while the weight of this term is $0.02-\beta$, and both are scaled down to $60\%$ with a cosine schedule until update $50{,}000$; $\gamma=0.1$ is warmed up over $1{,}000$ updates.

\paragraph{Policy.}
There are $16$ plan queries, $64$ future queries and $32$ action queries. The future queries are hidden from the action queries by the attention mask in every layer, so the plan is the only predictive signal the action head can read. The action head is a 16-layer DiT that integrates the flow in four steps. Learning rates are $10^{-5}$ for the VLM, $10^{-4}$ for the action head and $3\cdot10^{-6}$ for the world model.

\paragraph{In-context learning.}
The indexer is a two-layer MLP on the flattened plan ($16\times512\rightarrow256\rightarrow128$) with cosine logits at temperature $0.1$. Its weak supervision uses the annotated fine subtasks (frames of the same subtask are positives, with reviewed event intervals such as grasps as a stricter positive set), a return-cycle loss in both directions, and a soft-DTW path loss. The read-out window uses offsets $\Omega=\{-7,0,7,14\}$ frames around the matched stage; windows outside the demonstration are masked. The memory adapter has two cross-attention blocks with $16$ learnable queries over the $4\times64$ future-query tokens, and its $16$ output tokens have the width of the VLM. On LIBERO, the plans and content features used as keys and values are computed by a frozen copy of the policy that is refreshed every $1{,}000$ updates, so the alignment losses train the indexer but not the plan. The in-context components are added to a policy trained for $30$K updates without demonstrations: the indexer is warmed up for $2{,}000$ updates, and the policy is trained for $10$K further updates at a global batch of $256$, with a same-task demonstration from a different training episode attached to $80\%$ of the samples and zero memory for the rest. Demonstrations and test cases never share episodes.

\section{Imitator Game implementation}
\label{sec:suppl_imitator}
The Imitator Game instance uses the same queries on $\pi_{0.5}$~\cite{pi05} as the real-robot instance below, and $\lambda = 0.05$. The policy is initialised from the kai0 weights~\cite{kai0} and pretrained for $20$K updates on C15 and $50$K on C45, at a global batch of $256$ on four A100 GPUs; the in-context branch is switched on after $5$K updates. For P+FT the pretrained policy is fine-tuned for $10$K updates with all other settings unchanged.

\paragraph{Alignment loss.}
Let $q_i = g(Z_i)$, $i=1,\dots,n$, be the indexed plans of $n$ frames of a robot episode and $h_j = g(Z^{\mathcal{D}}_j)$, $j=1,\dots,m$, those of a human video of the same task, both of unit norm, and $s_{ij} = q_i^{\top} h_j / \tau$ with $\tau = 0.1$. The stage term is
\begin{equation}
    \mathcal{L}_{\mathrm{stage}} = -\frac{1}{|\mathcal{I}|}\sum_{i\in\mathcal{I}} \log \frac{\sum_{j\in\mathcal{P}_i} e^{s_{ij}}}{\sum_{j\in\mathcal{A}_i} e^{s_{ij}}} ,
\end{equation}
where $\mathcal{A}_i$ are the demonstration frames with a reviewed label, $\mathcal{P}_i\subseteq\mathcal{A}_i$ those in the interval reviewed as corresponding to the subtask of frame $i$, and $\mathcal{I}$ the frames for which $\mathcal{P}_i$ is not empty; frames at equal normalised time are not treated as positives. With $P = \mathrm{softmax}_j(s_{ij})$ and $Q = \mathrm{softmax}_i(s_{ij})$, the return distributions are $R = PQ^{\top}$ on the robot side and $R' = Q^{\top}P$ on the demonstration side, and the cycle term is
\begin{equation}
    \mathcal{L}_{\mathrm{cycle}} = \frac{1}{2n}\sum_{i,i'} R_{ii'}\,\big[\,|t_i - t_{i'}| - \epsilon\,\big]_+^2 + \frac{1}{2m}\sum_{j,j'} R'_{jj'}\,\big[\,|t'_j - t'_{j'}| - \epsilon'\,\big]_+^2 ,
\end{equation}
where $t$ and $t'$ are the times of the frames within their own videos and $\epsilon$, $\epsilon'$ tolerances; times are only compared within one video. The order term is the soft-DTW value~\cite{softdtw} of the cost $D_{ij} = 2 - 2\,q_i^{\top}h_j$ with smoothing $\gamma = 0.1$, debiased by the two self-alignments,
\begin{equation}
    \mathcal{L}_{\mathrm{order}} = \tfrac{1}{n}\Big[\mathrm{sdtw}_\gamma\big(D(q,h)\big) - \tfrac12\,\mathrm{sdtw}_\gamma\big(D(q,q)\big) - \tfrac12\,\mathrm{sdtw}_\gamma\big(D(h,h)\big)\Big] ,
\end{equation}
and $\mathcal{L}_{\mathrm{align}} = \mathcal{L}_{\mathrm{stage}} + \mathcal{L}_{\mathrm{cycle}} + \omega\,\mathcal{L}_{\mathrm{order}}$ with $\omega = 0.1$. In both the C15 and the C45 runs the loss is computed on $4$ episode pairs of $8$ frames each per update, on plans produced by the policy being trained for both the robot and the human frames, and its weight $\lambda$ is ramped up over $250$ updates.

\section{Real-robot implementation}
\label{sec:suppl_real}
\paragraph{Instantiation on $\pi_{0.5}$.}
We build on the JAX implementation of $\pi_{0.5}$~\cite{pi05} in kai0~\cite{kai0} and keep all of its parameters. After the image and language tokens of its VLM prefix we append $16$ plan queries, $16$ memory tokens and $64$ future queries; as in the Qwen3-VL instance, the plan queries read only the observation and the preceding plan queries, the memory tokens read the observation and the plan, and the action expert reads the observation, the plan and the memory but not the future queries. The plan is projected to $16\times512$ for the world model, and the future queries are decoded to the V-JEPA~2.1 feature grid of the frame a fixed offset ahead. The proprioceptive state is discretised into the language prompt as in $\pi_{0.5}$, and the $14$-D actions are zero-padded to the $32$-D action space of the checkpoint; the flow loss is computed on the $14$ valid dimensions only.

\paragraph{Data and training.}
The $100$ demonstrations are recorded in the training setup with the top camera and two wrist cameras, resized to $256\times256$, and each carries reviewed subtask segments, from which the $K=4$ checkpoints are computed as in \cref{sec:latent_plans}. Actions are predicted in chunks of $16$ steps. All parameters are fine-tuned for $20$K updates at a global batch of $256$, with a linear warm-up of $1{,}250$ updates followed by decay to the end of training. The in-context branch is switched on after $5$K updates: from then on, a demonstration recorded in a different session (a different \texttt{source\_group}) is attached to $80\%$ of the samples, and zero memory to the rest. On this task the stage is located by the cosine similarity between the flattened plans of the robot and of the demonstration, without a learned indexer; the demonstration plans and content features are cached in a bank computed by the current policy and refreshed every $2{,}500$ updates. The read-out window uses offsets $\{-16,0,16,32\}$ frames, \ie one chunk before, at, and one and two chunks after the matched stage. At test time the demonstration (one episode from the training setup, or one recorded in the deployment setup and excluded from training) is encoded once by the deployed policy, and the stage is matched by cosine similarity at every chunk.

\section{Held-out tasks of RoboTwin~2.0}
\label{sec:suppl_robotwin}
RoboTwin~2.0~\cite{robotwin2} is a bimanual benchmark of 50 tasks. We follow the split of Zero-WAM~\cite{zerowam}, which trains on 43 of them and evaluates on the other seven, where the policy sees the task only through a human video; the videos are those of the HumanGen Simulation ICL set released with Zero-WAM. \method reaches $42.3\%$ on the seven held-out tasks (\cref{tab:robotwin_heldout}), $+20.3$ points over the kai0 baseline~\cite{kai0} it is initialised from and $+5.9$ over Zero-WAM trained without its large-scale pretraining on paired human--robot data, the strongest of the three methods that \cite{zerowam} reports in this setting (WAN-Action $11.0\%$, LingBot-VA $17.5\%$). We restrict the comparison to this setting because our policy uses no such pretraining either; the comparison with Zero-WAM remains one between complete systems, since it starts from a video model and ours from a bimanual action policy.

\begin{table}[!ht]
\centering
\small
\setlength{\tabcolsep}{14pt}
\renewcommand{\arraystretch}{1.05}
\caption{\textbf{Held-out tasks of RoboTwin~2.0}~\cite{robotwin2} under the split of Zero-WAM~\cite{zerowam}: policies are trained on 43 tasks and evaluated on the other seven, for which a human video of the task is available as context. Average success rate (\%) over the seven tasks; the first three rows are reported in~\cite{zerowam}, and no row uses large-scale pretraining on paired human--robot data. Our policy is initialised from the kai0 weights, so the kai0 row is its controlled baseline.}
\label{tab:robotwin_heldout}
\begin{tabular}{lc}
\toprule
Method & SR \\
\midrule
WAN-Action & 11.0 \\
LingBot-VA & 17.5 \\
Zero-WAM~\cite{zerowam} w/o pretraining & 36.4 \\
\midrule
\abl{kai0~\cite{kai0}}                  & \abl{22.0} \\
\rowcolor{oursrow}
\textbf{\method (ours)}                 & \textbf{42.3} \\
\bottomrule
\end{tabular}
\end{table}

\paragraph{Training.}
The policy is the $\pi_{0.5}$ instance of \cref{sec:suppl_real}, initialised from the kai0 weights~\cite{kai0}. It is trained with the learning-rate schedule of a $40$K-update run and we report the checkpoint at $20$K updates. For the first $5$K updates no demonstration is given and the policy is trained on robot actions, the subtask plan and the future-query prediction, with the plan regressed onto the teacher plan for the first $625$ updates and the world model kept fixed until update $1{,}250$. From $5$K to $20$K updates the same supervision is kept, a human video is attached to $80\%$ of the samples and zero memory to the rest, and the optimiser state is carried over. The global batch is $128$ with seed $42$; training ran on four A100 GPUs up to update $10{,}565$ and on eight from then on.

\section{Additional ablations}
\label{sec:suppl_abl}
\begin{figure}[!ht]
    \centering
    \includegraphics[width=.62\linewidth]{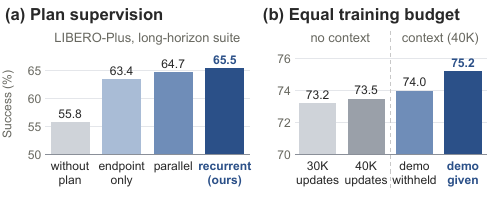}
    \caption{\textbf{Additional ablations on LIBERO-Plus.} (a) Long-horizon suite: the policy without the plan, and with a plan supervised by the subtask endpoint only, by the four checkpoints predicted in parallel, or by their recurrent prediction (ours). (b) Average of the policy trained without context for 30K or 40K updates, and of the context-trained checkpoint (40K) with the demonstration withheld or given.}
    \label{fig:suppl_abl}
\end{figure}
\paragraph{How the plan is supervised.}
\Cref{fig:abl}a removes the plan from the policy as a whole. On the long-horizon suite of LIBERO-Plus we also vary how the plan is supervised (\cref{fig:suppl_abl}a), with a re-trained run of our policy that reaches $65.5\%$ against $55.8\%$ without the plan; the run of \cref{tab:liberoplus} reaches $65.3\%$ on this suite, and $79.4\%$, $68.9\%$ and $69.4\%$ on the spatial, object and goal suites. Predicting only the subtask endpoint gives $63.4\%$ and predicting the four checkpoints in parallel $64.7\%$, against $65.5\%$ for the recurrent prediction of \cref{eq:rollout}. Having the plan thus matters more than how it is supervised; the order agrees with the design, although the margins are small for single runs, and the other suites were not run.

\paragraph{Training with context at an equal budget.}
The in-context checkpoint of \cref{tab:liberoplus} receives $10$K more updates than the policy it is compared with. As a control, we train that policy without context for the same $40$K updates: its LIBERO-Plus average moves from $73.2\%$ to $73.5\%$ (\cref{fig:suppl_abl}b). At an equal budget, training with context therefore adds $0.5$ points with the demonstration withheld ($74.0\%$), and reading a demonstration at test time adds $1.2$ more ($75.2\%$).

\end{document}